\documentclass[letterpaper]{article} 
\usepackage[draft]{style_file}  
\usepackage{times}  
\usepackage{helvet}  
\usepackage{courier}  
\usepackage[hyphens]{url}  
\usepackage{graphicx} 
\usepackage{natbib}  
\usepackage{caption} 
\usepackage{algorithm}

\usepackage{booktabs}
\usepackage{xcolor}
\usepackage{bm}
\usepackage{multirow}
\usepackage[nointegrals]{wasysym}
\usepackage[T1]{fontenc}
\usepackage{amsmath}
\usepackage{amssymb}
\usepackage{pifont}
\newcommand{\Plus}{\mathord{\text{\ding{58}}}}
\newcommand{\etal}{\textit{et. al}}
\usepackage{algpseudocode}
\usepackage{subcaption}

\definecolor{B1}{HTML}{1f77b4} 
\definecolor{B2}{HTML}{ff7f0e} 
\definecolor{SAL}{HTML}{9467bd} 
\definecolor{SAL_One}{HTML}{7f7f7f} 
\definecolor{TAIT}{HTML}{17becf} 
\definecolor{Avail}{HTML}{8c564b} 

\usepackage{newfloat}
\usepackage{listings}
\DeclareCaptionStyle{ruled}{labelfont=normalfont,labelsep=colon,strut=off} 
\floatstyle{ruled}
\newfloat{listing}{tb}{lst}{}
\floatname{listing}{Listing}
\title{Increasing Interpretability of Neural Networks \\By Approximating Human Visual Saliency} 
\author{Aidan Boyd,~Mohamed Trabelsi,~Huseyin Uzunalioglu,~Dan Kushnir}
\affiliations{Nokia Bell Labs\\Murray Hill\\NJ 07974, USA\\aidan.boyd@nokia-bell-labs.com}

\begin{document}

\maketitle

\vskip-10mm
\begin{figure*}[htb!]
\centering
\includegraphics[width=0.8\textwidth]{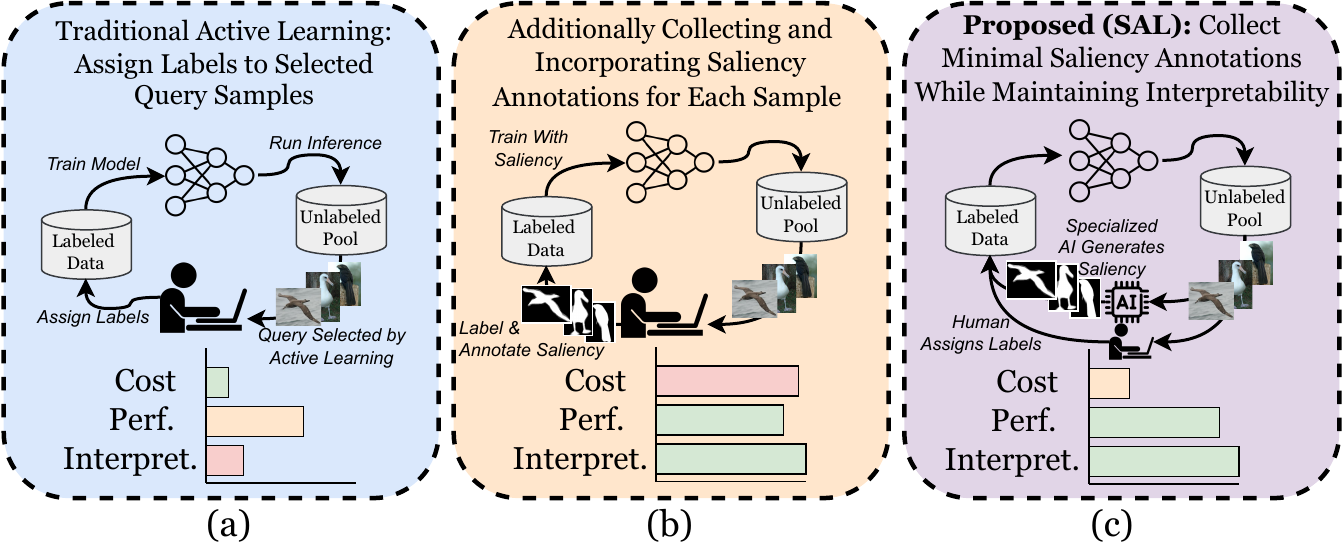}
\vspace{-0.8em}
  \caption{Overview of proposed approach. A traditional active learning pipeline is described in (a). In (b), this process is augmented to additionally collect saliency information about the images as well as the labels. These saliency annotations are then incorporated into the training process. In the proposed method, \textbf{S}aliency in \textbf{A}ctive \textbf{L}earning (SAL), human annotations are initially collected for a small number of iterations of active learning as in (b). After this, all future saliency annotation is delegated to an AI model specialized to produce high fidelity saliency maps (c), thus reducing overall human effort.}
  \label{fig:teaser}
\end{figure*}

\begin{abstract}

Understanding specifically where a model focuses on within an image is critical for human interpretability of the decision-making process.
Deep learning-based solutions are prone to learning coincidental correlations in training datasets, causing over-fitting and reducing the explainability. Recent advances have shown that guiding models to human-defined regions of saliency within individual images significantly increases performance and interpretability. Human-guided models also exhibit greater generalization capabilities, as coincidental dataset features are avoided.
Results show that models trained with saliency incorporation display an increase in interpretability of up to 30\% over models trained without saliency information.
The collection of this saliency information, however, can be costly, laborious and in some cases infeasible.  
To address this limitation, we propose a combination strategy of saliency incorporation and active learning to reduce the human annotation data required by 80\% while maintaining the interpretability and performance increase from human saliency. 
Extensive experimentation outlines the effectiveness of the proposed approach across five public datasets and six active learning criteria.

\end{abstract}

\section{Introduction}

Training human interpretable artificial intelligence (AI) models is vital to ensuring transparency, fostering trust, and enabling users to both understand and validate the AI decision-making process. High interpretability leads to more responsible and ethical applications of artificial intelligence.
Additionally, AI interpretability is becoming essential from a regulatory perspective because it addresses legal and ethical standards set by frameworks such as the EU's General Data Protection Regulation (GDPR) \cite{GDPR2016a}, mandating transparency in automated decision-making, the USA Algorithmic Accountability Act \cite{USAlgAcctAct}, aiming to ensure fairness and accountability in AI systems, and the EU AI Act \cite{EU_AI_Act}, designed to enforce safety, transparency, and accountability. These regulations, among others \cite{Secretariat_2017,ausAIethics,Docs_Italia}, collectively highlight the necessity for AI decisions to be understandable to humans to enable scrutiny and compliance with principles of equity, transparency, and user trust.

Modern neural networks (NN) have shown remarkable performance on many computer vision tasks including image classification \cite{deng2009imagenet,rawat2017deep}, object detection \cite{liu2020deep}, face recognition \cite{guo2019survey,wang2021deep,masi2018deep}, medical image analysis \cite{litjens2017survey} and biometrics \cite{sundararajan2018deep}. 
However, these models can lack interpretability because their internal structures involve multiple layers of complex computations and non-linear transformations, obscuring the path from input to output. This makes it difficult to decipher how individual image features are used in the decision-making process, contributing to their ``black-box'' nature. As such, NNs are susceptible to learning spurious features \cite{hovy2015tagging, pmlr-v80-hashimoto18a, pmlr-v139-zhou21g,buolamwini2018gender}, \textit{i.e.} features in the training data that are only coincidentally correlated with class labels such as the background color, position in the image or even features imperceptible to humans. These spurious features (or \textit{dataset biases}) drastically reduce the explainability of trained models. 
Thus, training models aligned with human perception is crucial for trustworthiness and interpretability.

Such alignment can be achieved by directly incorporating human saliency into the training process \cite{fel2022harmonizing,linsley2018learning}.
This approach simplifies the AI's decision-making process by prioritizing human-defined visually salient elements, therefore making the model's inference more transparent and understandable. By mimicking human visual attention, it bridges the gap between complex AI algorithms and intuitive human understanding, improving trust and clarity in how models analyze and interpret images. Additionally, by focusing on human saliency, the model is deterred from overfitting to spuriously correlated image features, ensuring that the learning process is grounded in genuinely relevant visual cues. Saliency in this work refers to the areas within an image that are useful for humans in making a classification decision. Interpretability refers to how understandable and aligned the decision making process of an AI model is to a human, \textit{i.e.} how close it is to human saliency.

However, a significant challenge with integrating human saliency into AI is the high cost associated with collecting human saliency data. This process often requires extensive eye-tracking studies \cite{Czajka_WACV_2019} or manual annotations by human participants \cite{fel2022harmonizing,boyd2021cyborg,boyd2023human,boyd2022human} to identify which regions within images are salient. Such methods are not only time-consuming but also may require specialized equipment and domain expertise, making the acquisition of large-scale, high-quality human saliency datasets an expensive, and in some cases infeasible endeavor. 
Such limitations highlight the importance of exploring efficient alternatives, such as active learning, which can significantly reduce the amount of labeled data required for training robust models.


Active learning is a semi-supervised machine learning approach that strategically selects the most informative and valuable data points from a pool of unlabeled images for manual labeling (shown in Fig. \ref{fig:teaser}(a)) \cite{ren2021survey}. The core idea behind this technique is to enable the learning algorithm to drive the data annotation process by identifying which data points, once labeled, would most significantly improve the model's capabilities. This selection process often relies on uncertainty sampling (refinement) \cite{nguyen2022measure}, where the algorithm queries data points about which it is most uncertain, or other strategies such as diversity sampling (exploration) \cite{yang2015multi}, which seeks to choose a set of diverse and representative images from the dataset. 
Active learning is particularly advantageous in computer vision tasks where labeling large datasets can be prohibitively expensive and time-consuming \cite{kaushal2019learning}. 
In this work, we propose \textbf{S}aliency in \textbf{A}ctive \textbf{L}earning (SAL) that a) increases model interpretability using saliency incorporation while maintaining or improving classification performance, and b) largely reduces the amount of human saliency data required to achieve this increase using active learning principles.

\begin{figure*}[t]
  \begin{subfigure}[b]{1\textwidth}
  \centering
      \centering
      \includegraphics[width=0.8\textwidth]{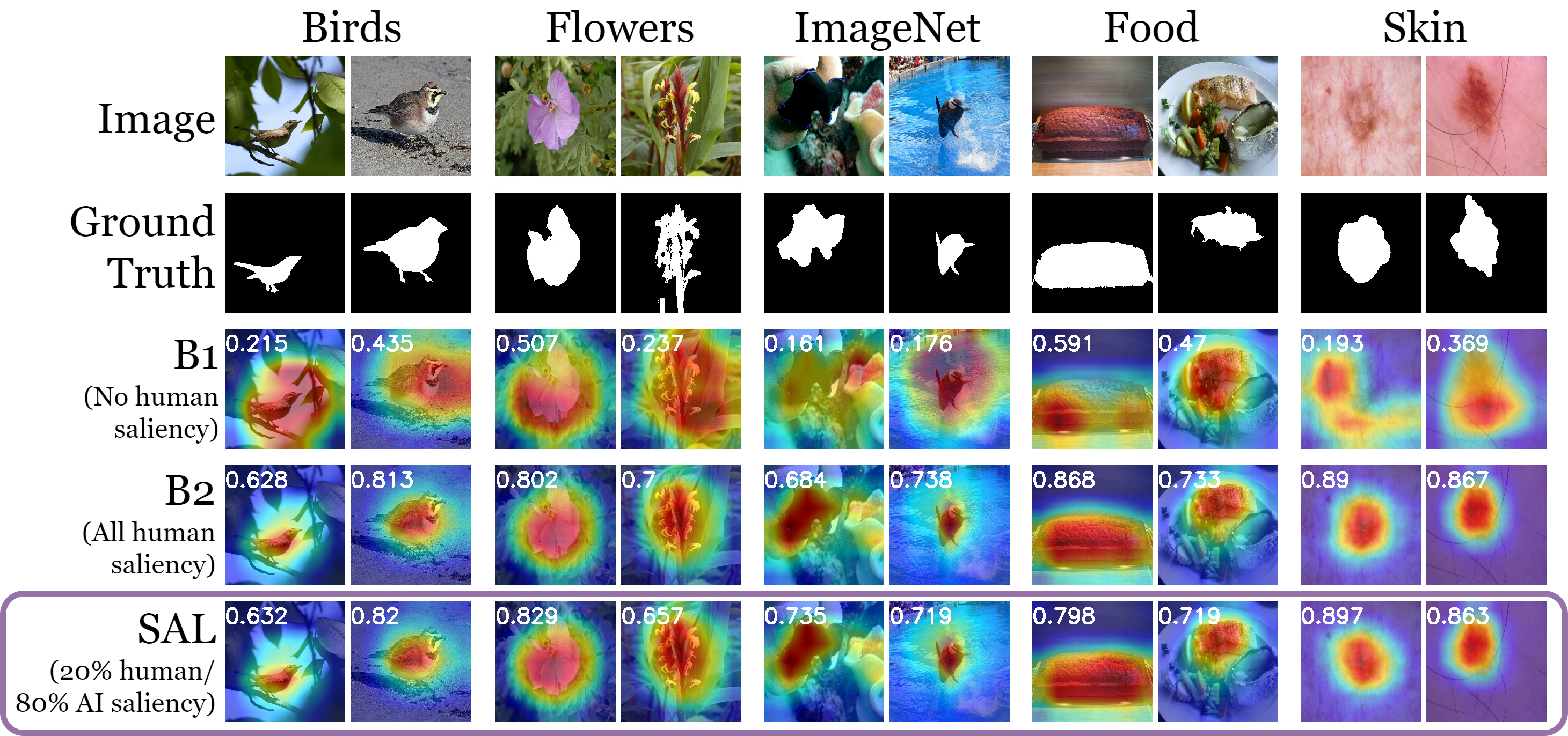}
  \end{subfigure} \vskip3mm
  \vspace{-1em}
  \caption{Two examples of model saliency for each of the five studied datasets. In each case, all three models (B1, B2, SAL) classified the image correctly. The DICE score with the ground truth is presented in the upper left corner of each heatmap. Models trained without saliency (B1) focus more on background and spurious features for classification. The model saliency of B2 and SAL are similar, with SAL only requiring 20\% of the amount of human saliency. 
  }
  \label{fig:examples}
  \vspace{-1em}
\end{figure*}

More specifically, in the initial iterations of the active learning pipeline, humans provide saliency annotations for each of the samples returned by the active learning criteria (query samples). These saliency annotations are then incorporated into the training process using the method proposed by Boyd \etal \cite{boyd2021cyborg} (as described in Fig. \ref{fig:teaser}(b)). Once a small set of human saliency annotations have been collected, dictated in this work as 20\% of the entire dataset, the saliency annotation task is then delegated to a specialized AI model that is trained to generate highly accurate saliency maps (Fig. \ref{fig:teaser}(c)), while the humans supply labels. Both the specialized interpretable model and human guided models are continually updated using active learning. This vastly reduces the work involved in the annotation process, as collecting labels is significantly quicker than both labeling and supplying detailed saliency annotations \cite{vondrick2013efficiently,dang2022vessel}. 

Results from extensive experimentation show that SAL significantly increases model interpretability compared to models without any saliency incorporated with either no effect or a slightly positive effect on accuracy, and matches the performance of models trained with 5 times as much human saliency annotations. Examples detailing the effectiveness of SAL can be seen in Fig. \ref{fig:examples}. This approach is shown to be applicable to any active learning criteria and results are validated on five publicly available datasets with associated saliency maps. Additionally, SAL is successfully applied to both Convolutional Neural Networks (CNN), and Transformer-based architectures, further emphasizing the universality.
In summary, this paper is structured to answer the following research questions:
\begin{itemize}
    \item RQ1: Does incorporating saliency into the training process increase model interpretability? Is there an effect on classification performance?
    \item RQ2: Can human saliency information be substituted with machine generated saliency, thus reducing annotation effort?
    \item RQ3: Is SAL broadly applicable, in that it can be used across active learning criteria, model architectures, datasets and saliency probing methods? 
\end{itemize}

\section{Related Work}

\paragraph{Estimating Model Saliency:} 
\label{sec:salience-methods} 

When estimating the saliency of models, access to the internal mechanisms such as the feature maps, gradients and weights largely improves overall fidelity. The most popular of these so-called white-box approaches is Class Activation Mapping (CAM) \cite{zhou2016learning}. CAMs are generated by making a forward pass through the model to get the activations of the last convolutional layer. Using these activations as weights, a weighted sum of the feature maps of this last convolutional layer is created. The resulting heatmap represents the regions in the image that the model deems most salient. Advances such as Grad-CAM \cite{gradcam}, Grad-CAM++ \cite{gradcam++}, HiResCAM \cite{hirescam}, Score-CAM \cite{scorecam}, Ablation-CAM \cite{ablationcam}, or Eigen-CAM \cite{eigencam} aim to estimate more detailed saliency, but require more computational resources such as access to gradients or multiple forward passes.

Recent work on black-box explainers  (no access to model internals) include methods of evaluating their usefulness for humans \cite{Carmichael_ArXiv_2023} and increasing their robustness against adversarial attacks \cite{Carmichael_AAAI_2023}.
The most popular black-box methods for visual saliency randomly perturb input regions and observe the impact on the output \cite{petsiuk2018rise}. 
Due to the increased computational expense of black-box methods, and can only be employed on already trained models (post-hoc), this work focuses only on CAM as the method of saliency estimation. Experiments using other saliency probing methods are found in the supplementary materials.

\paragraph{Human Saliency-Guided Model Training:}    
Human perception can be captured by various means such as image/video annotations \cite{boyd2021cyborg}, eye-tracking \cite{boyd2023human,Czajka_WACV_2019}, reaction times \cite{huang2022measuring,grieggs2021measuring}, or even by playing games \cite{linsley2018learning}. Successful attempts of incorporating this human-collected information into the training process include adding specialized components to the loss functions \cite{boyd2021cyborg, huang2022measuring}, augmenting training data \cite{boyd2022human}, pre-training the model to include saliency information \cite{crum2023mentor}, and the introduction of human perception-based regularization \cite{huang2022measuring, dulay2022using}. The CYBORG training strategy \cite{boyd2021cyborg} incorporates human guidance in the loss function by penalizing the divergence of the model's CAM from the human saliency provided as image annotations. Fel \etal  \cite{fel2022harmonizing} proposed a neural harmonizer to align image classification models and human visual strategies. The neural harmonizer computes the feature importance of a differentiable network (classification model) using gradient-based saliency of the network with respect to the input.

In a recent work by Crum \etal~\cite{crum2023teaching}, the authors show how manipulating the parameter that balances the classification component of the loss and the human saliency component in CYBORG loss can enable saliency generation on unannotated data. 
Saliency generation models (called \textit{teacher models}) are trained in a fully-supervised manner using human saliency annotations. They show that it is possible to leverage a small set of human annotations to create more accurate models by synthetic saliency generation. 
\textbf{Our work furthers from this as}, instead of a static saliency generation model trained only in the fully-supervised manner as in \cite{crum2023teaching}, we iteratively update the model with AI generated saliency in a semi-supervised manner using active learning. 

\section{Saliency Incorporation with CYBORG}
\label{sec:cyborg_loss}

While there are various means of incorporating saliency into the training procedure of CNNs such as guided-attention mechanisms \cite{linsley2018learning} and saliency-specific augmentations \cite{boyd2022human}, promising results have been demonstrated in loss-based saliency incorporation strategies \cite{boyd2021cyborg,fel2022harmonizing}. 

The CYBORG approach is a multi-objective loss strategy that combines the human saliency information attained through manual annotation (\textit{human saliency loss}) with classification performance (\textit{classification loss}). The human saliency loss directly compares the difference in salient regions between the model and humans during training, steering activations in the feature maps in the last convolutional layer to be aligned with human-defined regions of importance. The classification loss component ensures that the model learns the class labels of the images in a data-driven manner. 
To attain model saliency, the authors used the Class Activation Mapping (CAM) approach~\cite{zhou2016learning}, which represents the most simple and resource-efficient approach to model saliency probing. 

As detailed in \cite{boyd2021cyborg}, the CYBORG loss $ \mathcal{L}$ is defined as:
\vskip-3mm

\begin{equation}
\resizebox{1\hsize}{!}{$
\begin{split}
\mathcal{L} = \frac{1}{K}\sum_{k=1}^K\bm{1}_{y_k \in c} 
\Bigg[\underbrace{(1-\alpha)\mathcal{L}_s\big(\textbf{s}_k^{\text{(h)}},\textbf{s}_k^{\text{(m)}}\big)}_{\text{human saliency loss}} -\underbrace{\alpha\log p_{\text{m}}\big(y_k \in c\big)}_{\text{classification loss}}\Bigg]
\end{split}$}
\label{eqn:cyborg}
\end{equation}

\noindent
where $\mathcal{L}_s$ is a measure comparing model and human saliency maps, $y_k$ is a class label for the $k$-th sample, $\bm{1}$ is a class indicator function equal to $1$ when $y_k \in c$ (0 otherwise), $c$ is the class label, $K$ is the number of samples in a batch, $\alpha$ is a trade-off parameter weighting human- and model-based saliencies, $\textbf{s}_k^{\text{(h)}}$ is the human saliency for the $k$-th sample, and $\textbf{s}_k^{(\text{m})}$ is a class activation map-based model's saliency for the $k$-th sample. $\mathcal{L}_s$ is the sum of structural similarity (SSIM) and L1 distance.

\section{SAL: Saliency in Active Learning}
\label{sec:sal_description}

\begin{algorithm}[t]
\caption{Saliency in Active Learning (SAL)}\label{alg:SAL}
\begin{algorithmic}[1]

\State \noindent\textbf{Given:} unlabeled samples $X_{pool}$
\State \textbf{Collect:} labels and saliency annotations for initial subset $X_{l}$ from humans
\State \textbf{Set:} C, N \Comment{C = Change point, N = Num AL iterations}
\For{\textit{i} = 0; \textit{i} $\leq$ N; \textit{i}++}
\State \textbf{Train:} Model $M_{i}^{acc}$ using $X_{l}$ and saliency \\ \qquad incorporation ($\alpha=0.9$)
\State \textbf{Train:} Model $M_{i}^{interp}$ using $X_{l}$ and saliency\\ \qquad incorporation ($\alpha=0.1$)
\State \textbf{Infer:} On remaining pool ($X_{pool}$-$X_{l}$) using $M_{i}^{acc}$
\State \textbf{Select:} Query set ($Q_{i}$) using AL selection criteria
\If{i < C} 
    \State \textbf{Collect:} labels \& saliency for $Q_{i}$ from humans
\Else 
    \State \textbf{Collect:} labels for query set $Q_{i}$ from humans
    \State \textbf{Generate:} saliency for $Q_{i}$ using $M_{i}^{interp}$
\EndIf
\State \textbf{Add:} query set $Q_{i}$ with labels and annotations to $X_{l}$
\EndFor
\State \textbf{Test:} on test set using final $M^{acc}_{N}$

\end{algorithmic}
\end{algorithm}

In this section, we introduce SAL, a novel interpretable training approach based on \textbf{S}aliency incorporation in \textbf{A}ctive \textbf{L}earning.\footnote{Code, dataset splits and models found here: <anonymized>}
Initially, as shown in Fig. \ref{fig:teaser}(b), during the labeling stage of active learning, humans are additionally queried to supply saliency annotations detailing the regions most pertinent to their classification decision. This additional visual annotation step is costly and laborious, thus, must be reduced as much as possible. In SAL, human annotations are collected until 20\% of the total training set is annotated, from which point specialized AI models are delegated to perform this task of visual annotation while the human annotators supply labels. SAL is detailed in Algorithm~\ref{alg:SAL}.

By adjusting the balancing parameter ($\alpha$, Eq. \ref{eqn:cyborg}) in CYBORG loss \cite{boyd2021cyborg}, the emphasis is shifted between predicting the salient regions in an image and the classification of the image. Leveraging this flexibility, during each active learning iteration two distinct models are trained on the same set of currently labeled data; one is trained with the balance parameter heavily skewed towards classification performance ($\alpha$=0.9, Alg. \ref{alg:SAL}: $M^{acc}$), and the second auxiliary model is trained with the same parameter heavily skewed to interpretability ($\alpha$=0.1, Alg. \ref{alg:SAL}: $M^{interp}$). Model $M^{acc}$ then selects samples from the unlabeled pool set with the highest uncertainty for further labeling, after which $M^{interp}$ is employed to generate saliency maps for those selected samples. The AI generated masks from $M^{interp}$ are then used for saliency incorporation in the next round of model training. These masks represent high quality approximations of human saliency. $M^{acc}$ is used for final testing.
The set of all saliency maps used by SAL is a combination of both the initial human supplied annotations and iteratively generated AI saliency. 
SAL is detailed in Fig. \ref{fig:teaser}(c). The AL query size is set to be 5\% of of the total train set. Thus, for this work, $C=5$ and $N=20$ in Alg. \ref{alg:SAL}.

\section{Experimental Setup}

Baselines introduced in this section are designed to evaluate the effectiveness of SAL against the current state-of-the-art in saliency generation for classification tasks and to answer the research questions posed in the introduction.  
Six AL criteria are extensively studied representing exploration (CoreSet \cite{sener2017coreset}, Random Sampling), refinement (Least Confidence, Entropy Sampling \cite{settles2009active}) and a combination of both (Margin Sampling \cite{scheffer2001active}, BADGE \cite{ash2019badge}). 
Explanations of each algorithm and model training parameters can be found in the supplementary materials. 

\paragraph{Baseline 1 (B1):} 

The first and most simple training scenario is when no saliency information is incorporated into the training process. This baseline represents the standard active learning pipeline where actively selected query images are annotated with only a label at each iteration. Models are trained in a traditional way where only a classification loss (categorical cross-entropy) is utilized, as there is no saliency information available. B1 is detailed in Fig. \ref{fig:teaser}(a). \textbf{This represents the lower bounds of model interpretability}, as any features the model learns are directly from the training set without any human guidance.

\paragraph{Baseline 2 (B2):}

The second baseline scenario is the hypothetical situation where all images have their salient regions annotated. \textbf{This represents the upper bounds of what is possible given all saliency annotations available and incorporated during training using CYBORG}. The goal of SAL is to attain performance as close to this baseline as possible while reducing the amount of annotation data required. B2 is detailed in Fig. \ref{fig:teaser}(b). 

\paragraph{Teaching AI to Teach (TAIT):} 
Crum \etal \cite{crum2023teaching} proposed a novel method of synthesizing saliency for unannotated samples using AI.
These AI saliency generator \textit{teacher} models are trained in a fully-supervised manner using human annotations and CYBORG loss. In \cite{crum2023teaching}, the architecture used to generate the AI saliency was Xception \cite{chollet2017xception}, and the $\alpha$ parameter in CYBORG loss is set to 0.5. 
In this baseline, once all human annotation data is collected, the model trained to generate saliency is frozen, and used in that state for all remaining iterations of active learning. 
\textbf{This baseline determines the usefulness of active learning in SAL} as it explores whether the addition of AI generated saliency to the training set for the interpretable model adds value to the overall performance, or whether there is no more interpretability to be gained once human annotation stops.
This approach represents the state-of-the-art in automatic saliency generation for image classification and is thus the best comparison for SAL.

\noindent \textbf{Of note} is that we additionally attempted to employ the neural harmonizer proposed in Fel \etal \cite{fel2022harmonizing} as the human saliency incorporation method. Our endeavors did not yield the anticipated results as the models did not align with human saliency and classification performance was negatively impacted. This suggests a need for further exploration into the method's adaptation to the data-limited nature of active learning. 
Conversely, CYBORG was developed using small datasets, providing a more natural application to active learning. 

\paragraph{Ablation Study:}

To validate the dual model approach within SAL, two related variants are also investigated. 

The \textbf{SAL (Single)} variant \textit{assesses the usefulness of the specialized model trained for interpretability}. In this variant, there is no specialized model trained for interpretability (Alg. \ref{alg:SAL}: $M^{interp}$ is not trained), instead the AI generated saliency is supplied by the model that was trained for accuracy ($\alpha=0.9$, Alg. \ref{alg:SAL}: $M^{acc}$). 

The \textbf{No AI Saliency} variant \textit{determines the overall usefulness of generating AI saliency}. In this variant, after the collection of human annotations stops, saliency incorporation is only applied to samples with saliency annotations. For all newly collected samples with labels alone, only classification loss is used. Within a single batch, there may be samples with saliency annotations (thus the saliency based loss is applied to them, $\alpha=0.9$), and some with no saliency annotations (classification loss only).

\paragraph{Additional Experiments}

An experiment replacing the human annotations in $B2$ with automatically generated masks using the off-the-shelf Segment Anything Model (SAM)\cite{kirillov2023segment} is detailed in the supplementary materials. We selected the output mask with the highest \textit{predicted\_iou} as the segmentation. Results of this experiment show that off-the-shelf masks are not effective replacements for human annotations, but improve interpretability over the no-saliency setting. Thus, the initial effort investment of collecting a small number of annotations is worthwhile due to the significant increase in interpretability attained.

In this work we set the change point from human to machine saliency to 20\%. To examine the effect of this parameter two additional experiments are run; collecting only 5\% human annotations and 10\% human annotations. Results in the supplemental materials show SAL is highly effective with just 5\% human annotations, while performance increases with additional human annotations.

\section{Evaluation}

\begin{table}[t]
\centering
\caption{Number of classes and samples in used datasets.}
\begin{tabular}{|c||c|c|c|c|}
\hline
Dataset & Train & Val & Test & Classes  \\ \hline\hline
CUB-200 &  4,794  & 1,200 & 5,794  &  200 \\ \cline{1-5}
Flowers102 & 1,020   &  1,020   &   6,149  & 102   \\ \cline{1-5}
ImageNet-S &  6,433  & 2,757   &  12,419  & 919  \\ \cline{1-5}
HAM1000 & 4,893   &  2,092  &  3,030 &  7   \\ \cline{1-5}
Food201 & 6,244  &  2,684 & 2,286  &  99  \\ \cline{1-5}
\end{tabular}
\label{tab:dataset_breakdown}
\vspace{-1em}
\end{table}

\begin{figure*}[t]
      \centering
    \includegraphics[width=1\textwidth]{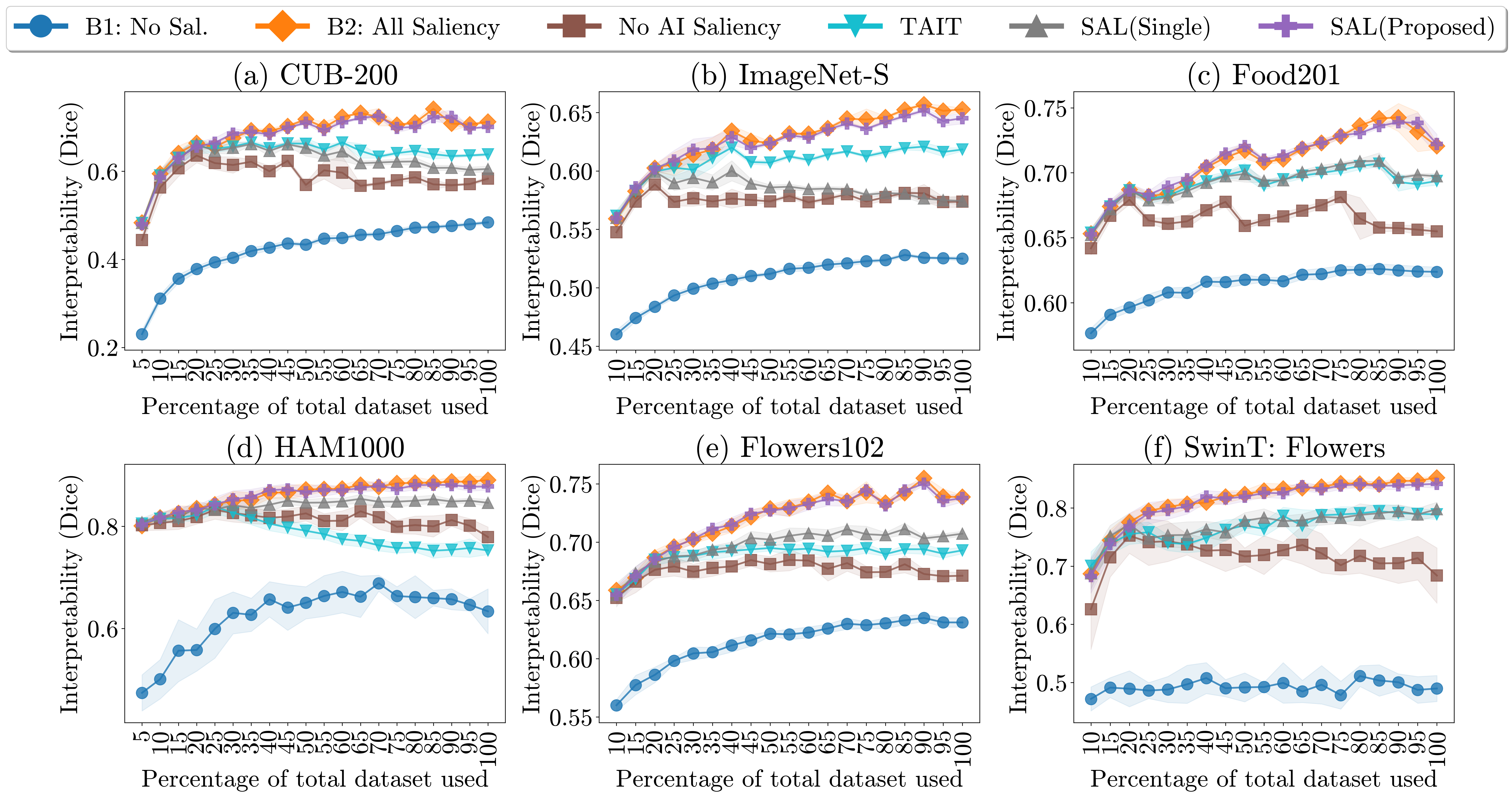}
 \vskip3mm
 \vspace{-1em}
  \caption{Learning curves comparing the overlap of model saliency trained under various scenarios with the ground truth on the test set for five datasets. In all cases the AL criteria was margin uncertainty. Plots (a)-(e) are ResNet50-based, while (f) shows the use of SAL with SwinTransformer. Aligning with the research questions in the introduction, results show the following: 1) models trained with saliency incorporated (\textcolor{B2}{B2}/\textcolor{SAL}{SAL}) have significantly higher overlap/interpretability than those trained without (\textcolor{B1}{B1}), 2) \textcolor{SAL}{SAL} effectively replicates the performance of \textcolor{B2}{B2} with 80\% fewer human annotations, and 3) the same trends can be seen across all five datasets. Each learning curve shows the mean of 8 AL runs, with the shaded area representing $\pm1\sigma$.}
  \label{fig:test_overlaps}
  \vspace{-1em}
\end{figure*}

\subsection{Metrics}

After each iteration of active learning, the trained model is tested to show performance gain as the number of labeled training samples grows. The plot showing this performance across iterations is called the learning curve (e.g. Fig. \ref{fig:test_overlaps}). The area beneath this learning curve (or area under the budget curve \cite{zhan2021comparative}) is a numerical representation of the models performance across all active learning iterations, with higher values representing better performance. 

The main metric used to evaluate classification performance is accuracy. Thus, the area under the learning curve representing the accuracy is called $AULC_{acc}$. The interpretability performance metric of the study is the Dice similarity coefficient (also known as F1 Score) \cite{dice1945measures}, which measures the relative overlap of the predicted and ground-truth masks. 
To convert the model CAMs to binary masks, the top $N$ highest value pixels are set to 1, and the rest are set to 0, where $N$ is the number of positive pixels in the ground truth. When no ground truth is available, such as when autonomously generating saliency, a threshold of 0.5 is applied to the model CAM to generate the binary mask.  The area under the learning curve representing the Dice and therefore the interpretability is called $AULC_{interp.}$.

\subsection{Datasets}

We evaluate SAL on five public datasets. 
These datasets were selected as all images have a corresponding saliency annotation. 
Tab. \ref{tab:dataset_breakdown} outlines the number of images and classes in the train, validation and test sets for each dataset. 

\textit{CUB-200 (Birds) \cite{WahCUB_200_2011}} is a fine-grained classification dataset with 200 categories of mostly North American birds. 
\textit{Flowers102 (Flowers) \cite{nilsback2008automated}} is a dataset consisting of 102 flower categories commonly occurring in the United Kingdom. Images have large scale, pose and light variations. 
\textit{ImageNet-S (ImageNet-S) \cite{gao2022luss}} is an annotated subset of ImageNet \cite{deng2009imagenet} adapted for semantic segmentation. 
\textit{Food201 (Food) \cite{bossard14}} is derived from the Food101 food classification dataset. Images are assigned one class per image (the primary class from Food101). 
\textit{HAM1000 (Skin) \cite{skin_dataset_ham1000}} is a dataset intended to train models for automated diagnosis of pigmented skin lesions across 7 categories.

\subsection{Results}

\begin{figure}[t]
  \begin{subfigure}[b]{1\columnwidth}
  \centering
      \centering
      \includegraphics[width=1\columnwidth]{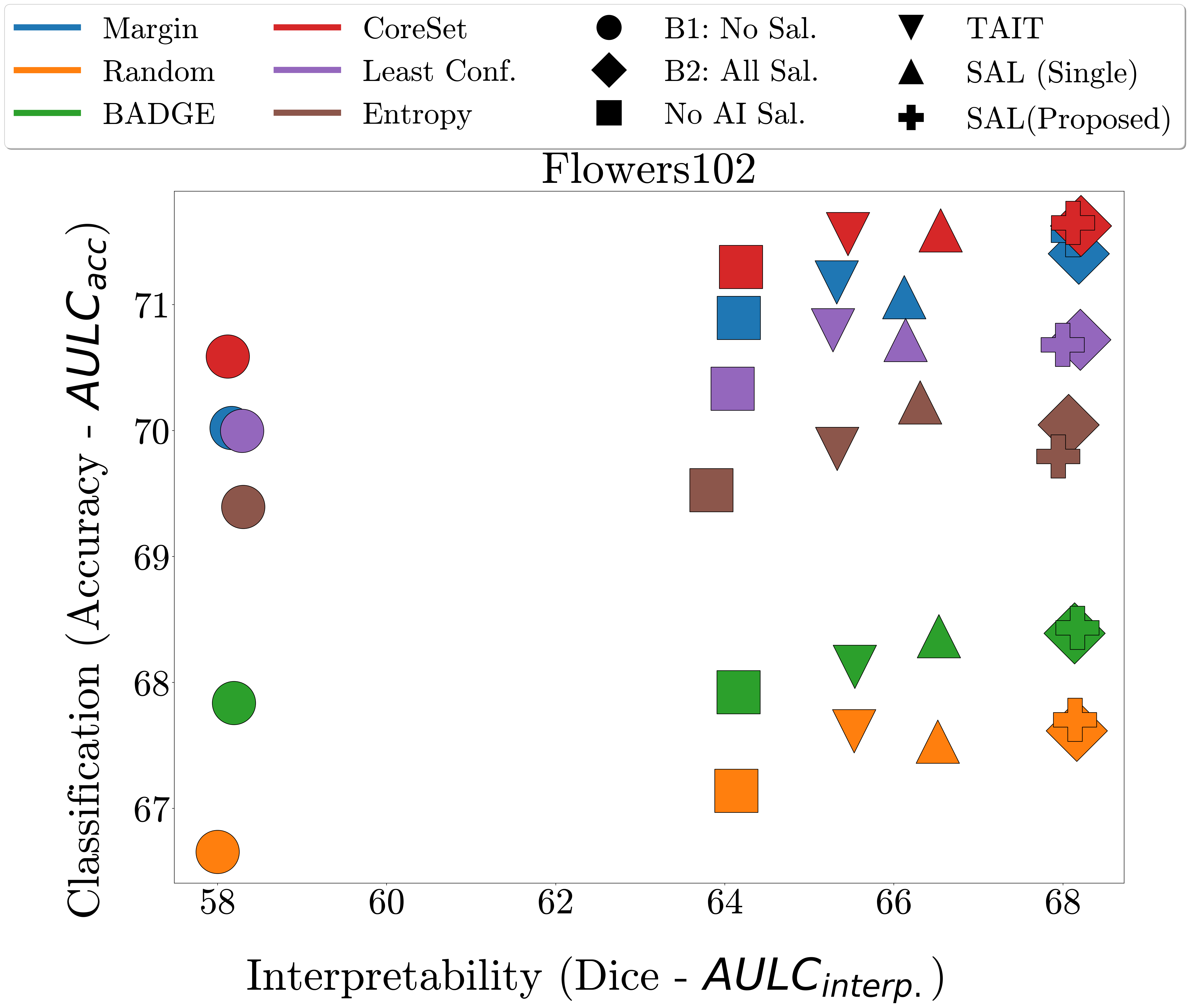}
  \end{subfigure} \vskip3mm
  \vspace{-1.6em}
  \caption{Plot details the classification performance over the interpretability. The shapes represent the training approach used, and the colors represent the active learning selection criteria used. Results show: 1) SAL ($\Plus$) matches the classification performance and interpretability of models trained with full saliency ($\blacklozenge$) across all AL criteria, and 2) SAL increases performance over TAIT baseline ($\blacktriangledown$), showing the effectiveness of combining saliency incorporation with active learning. Each point is the mean of 8 independent runs.}
  \label{fig:main_results}
  \vspace{-1em}
\end{figure}

\paragraph{\textbf{RQ1: Does incorporating saliency into the training process increase model interpretability? Is there an effect on classification performance?}}

This question is answered by comparing the results of Baseline 1 (B1) and Baseline 2 (B2). The only difference between these baselines is that in B2, human annotations are available for all images and incorporated into the training process. Looking at both Fig. \ref{fig:test_overlaps} (B1=\textcolor{B1}{Blue}, B2=\textcolor{B2}{Orange}) \& Fig. \ref{fig:main_results} (B1=\newmoon, B2=$\blacklozenge$), it is immediately evident that there is a large gain in interpretability when saliency is incorporated into training. In addition, as illustrated in Fig. \ref{fig:main_results} (B1=\newmoon, B2=$\blacklozenge$), the classification performance is increased. \textbf{Accuracy learning curves are supplied in the supplementary materials.}
As per Fig. \ref{fig:test_overlaps}, this gain in interpretability is visible for all five datasets and for both ResNet50 (a-e) and SwinTransformer (f). 
Before incorporation, the interpretability of the SwinT is lower than ResNet50, however, after saliency incorporation, the interpretability peaks that of ResNet. This details how transformer based architectures can be effectively guided towards human-defined image features.  

Thus, to conclude and answer RQ1: \textbf{the incorporation of human saliency into the training process significantly increases the interpretability of models, and has a positive effect on classification performance}.

\paragraph{\textbf{RQ2: Can human saliency information be substituted with machine generated saliency, thus reducing annotation effort?}}

The incorporation of human saliency into the training process during active learning demonstrably increases interpretability, the next question is whether this human saliency can be approximated by an AI model. To answer this, we propose SAL, an active learning-based approach that enables the effective generation of image saliency annotations. 
By examining the plots in Fig. \ref{fig:main_results}, when comparing SAL (Fig. \ref{fig:main_results} - $\Plus$) to B2 (Fig. \ref{fig:main_results} - $\blacklozenge$), the performance both in terms of classification accuracy and interpretability are largely comparable.
This similar performance is achieved with up to 80\% less human annotation data in SAL. Interestingly, in all cases in Fig. \ref{fig:test_overlaps}, SAL  (\textcolor{SAL}{Purple}) shows a positive slope as the training dataset increases. This is the same as B2. However, for SAL, no more human annotations are collected after attaining a budget of 20\%. This means that the interpretability models ($M^{interp}$) used in SAL can replicate human annotators so effectively that the models can increase knowledge on the domain without explicit human annotations. This comes with no negative effect on accuracy, avoiding the accuracy/interpretability trade-off.

In addition, this work investigates one state-of-the-art method and two variants of SAL to validate the design choices. The results from this ablation study confirms the following in relation to SAL (Fig. \ref{fig:test_overlaps} - \textcolor{SAL}{Purple}/Fig. \ref{fig:main_results} - $\Plus$): 
(1) Generating saliency for unannotated images using AI is significantly more effective than only using saliency incorporation on samples that have an associated human annotation (SAL vs. No AI Saliency, Fig. \ref{fig:test_overlaps} - \textcolor{Avail}{Brown}/Fig. \ref{fig:main_results} - $\blacksquare$). 
(2) Generating AI saliency using a specialized model tuned for interpretability ($M^{interp}$) yields saliency maps that are much more effective at guiding the model towards salient regions compared to using a model tuned for accuracy ($M^{acc}$) to generate saliency (SAL vs. SAL (Single), Fig. \ref{fig:test_overlaps} - \textcolor{SAL_One}{Gray}/Fig. \ref{fig:main_results} - $\blacktriangle$.
(3) Continually training the interpretability model ($M^{interp}$) at each step of active learning is significantly better than freezing the interpretability model once human annotation collection ceases (SAL vs. TAIT baseline, Fig. \ref{fig:test_overlaps} - \textcolor{TAIT}{Light Blue}/Fig. \ref{fig:main_results} - $\blacktriangledown$).
In conclusion, SAL's dual model, continually updated approach produces high fidelity AI saliency which can be incorporated into training to increase model interpretability.

To answer RQ2: we propose SAL, which when given only a small subset of human saliency annotations, can match classification performance and interpretability of models trained on a full set of images with all available human saliency incorporated. This reinforces that using SAL, \textbf{human saliency information can be substituted with machine generated saliency after minimal annotation effort.}

\paragraph{\textbf{RQ3: Can SAL be applied universally?}}

In RQ2, it was shown that the SAL approach can effectively replace human annotations, thus reducing overall annotation efforts. To reinforce utility, SAL has been successfully applied to: 
(1) Five publicly available datasets with saliency annotations are employed for each experimental setup, showing that results are domain agnostic - Fig. \ref{fig:test_overlaps}, Fig. \ref{fig:main_results}. 
(2) Six active learning criteria on each dataset (incl. two state-of-the-art approaches~\cite{ash2019badge,sener2017coreset}). These include refinement techniques, exploratory criteria and combinations of both. SAL does not rely on a single sample selection strategy, and even works when samples are randomly selected from the unlabeled set  - Fig. \ref{fig:test_overlaps}, Fig. \ref{fig:main_results}. \textbf{Importantly}, the goal of this work was not to discover the optimal active learning criteria for SAL, but that it can be used independently of the criteria.
(3) Both CNN-based (ResNet50~\cite{resnet}) and Transformer-based architectures (SwinTransformer~\cite{liu2021swin}, figures for SwinT on additional datasets in supplementary materials). 
(4) Four methods to estimate model saliency (CAM, GradCAM, GradCAM++, HiResCAM - see supplementary materials). In all cases similar trends as for CAM (Fig. \ref{fig:test_overlaps}, Fig. \ref{fig:main_results}) are observed.

Thus, to answer RQ3: \textbf{results show that SAL can be effectively applied across datasets, active learning criteria, model saliency probing methods and architectures.}

\section{Conclusions}

Training interpretable AI models is paramount for increasing trust, guaranteeing accountability, and upholding ethical standards within AI applications. 
A large increase in interpretability can be achieved by incorporating human saliency into the training process. However, the acquisition of this human saliency may be expensive, thus a reduction in human annotation time and effort is crucial.

This work addresses this labor-intensive nature of manual saliency labeling. 
We propose SAL, a novel approach combining saliency incorporation with active learning that significantly increases both the interpretability of models and classification performance. 
Saliency incorporation increases model interpretability by up to 30\%.
Results from this paper show that the proposed method can match this interpretability increase using 80\% less saliency annotations.
Experimental results also show the robustness of the approach, as the same trends are demonstrated across five public datasets, six different active learning criteria, both CNNs and Transformer based architectures and four saliency probing methods.
SAL is inherently applicable to large-scale datasets, as initially a small subset of human annotations is collected, which is then used to approximate saliency for any number of future samples. 

\bibliography{main}

\appendix

\section{Active Learning Specific Details}

In this work, the AL query size is set to be 5\% of the size of the total initial pool size. Thus, the exact query size varies from dataset to dataset. The starting point is also different for various datasets. To ensure at least one sample from each class is present in the initial labeled set, the starting point for ImageNet and Flowers was 10\% of the total data (each class only has 10 samples in the training set). For Birds and Skin, there are many more images per class, thus we set the starting point of these to be 5\% of the data. The starting point for the Food dataset was also 10\%. The selected datasets represent cases where there are many images per class in the training set (Birds, Food, Skin) and cases where there are few images per class (Flowers, ImageNet-S). 

\subsection{Descriptions of Active Learning Criteria}

\paragraph{Random Sampling:} Randomly select samples from the remaining pool set for the next query. This can be considered an exploratory criteria because it uniformly samples across the remaining samples, building a more expansive knowledge of the space.

\paragraph{Margin Sampling:} Samples are selected that have the smallest difference between the top two most confident predictions. Practically, this means samples with high confusion between two classes. In the early stages when prediction values are low, margin sampling can be considered an exploratory algorithm, however, as the models mature it turns into a refinement algorithm as it is explicitly defining the decision boundary between the top two predicted classes.

\paragraph{Least Confidence:} Samples are selected with the largest difference between the most confident prediction and 100\% confidence. These are samples which model does not have high confidence in. This is a refinement approach as it explicitly define decision boundaries for uncertain samples.

\paragraph{Entropy Sampling:} Samples are selected with high uncertainty or entropy in the set of class predictions. High entropy indicates that the model is uncertain about the correct class assignment for a particular instance, making it a good candidate for further labeling to improve the model's performance. The rationale behind entropy sampling is that instances with high entropy are considered more informative as the model is less confident about its predictions. 

\paragraph{Batch Active learning by Diverse Gradient Embeddings
(BADGE) \cite{ash2019badge}:} Point groups that are disparate and high magnitude when represented in a hallucinated gradient space are sampled, so that the prediction uncertainty of the model and the diversity of the samples are simultaneously considered in a batch. The goal of BADGE is to achieve an automatic balance between the uncertainty and sample diversity without the need for hyperparameter optimization.

\paragraph{Core-Set \cite{sener2017coreset}:} By attempting to find a core-set of samples, this approach aims to find a small subset given a large labeled dataset such that a model learned over the small subset is competitive over the whole dataset. Because Core-Set attempts to cover the entire sample space in its core set in an optimal way, it is an exploratory criteria.

\section{Limitations of this work}

We acknowledge there are limitations to this study. The first limitation we acknowledge is the detection of the point at which to change from collecting human saliency to generating AI-based saliency. In this work we set this to 20\%. Included in this supplementary materials are experiments using 5\% and 10\%, showing SAL still performs well at these points. Future work consists of dynamically locating this change point to be that at which the model trained for interpretability has sufficient knowledge of the domain such that the switch to AI saliency is as seamless as possible. 

Additionally, the primary proposal in SAL is to train a second model for interpretability. We acknowledge that this doubles training resources required. However, we believe the boost in interpretability justifies the additional resource requirements. 

We claim there is a large reduction in human effort by switching to labeling only instead of labeling and annotating. However, we do not provide quantitative proof of this in the form of time savings. This is because none of the datasets employed have this information available. It is the hope of the authors that the reader can intuitively see how assigning a label to an image is significantly quicker and easier than labeling and intricately annotating an image.

Finally, it is assumed that no saliency masks are available for the test set, meaning the model is not guided during testing. With the emergence of technologies such as the Segment Anything Model (SAM) \cite{Kirillov_2023_ICCV} and various Salient Object Detection methods \cite{borji2019salient}, it may be possible to generate saliency for the testing images during inference. We run an experiment using SAM in place of human saliency in Section \ref{sec:ots} of this supplementary materials. Results show that SAM generated saliency is preferable to no saliency incorporation, but using SAL with an initial set of human annotations is better.

However, the SAL framework is adaptable, and the proposed interpretability model can be modified to be any model the engineer selects. The main goal of SAL is to show that training two models in an active learning framework can significantly reduce annotation overhead. In this work, we show how the same model architecture can be modified to fulfill both purposes using CYBORG loss, these model architectures may be different though.
As mentioned in the main text, it is part of future work to investigate whether generalized segmentation models and salient object detection models can be directly applied to the SAL framework, replacing the current interpretability model.

\section{Hardware Resources}

\begin{itemize}
    \item \textbf{CPU:} AMD EPYC 7513 32-Core Processor.
    \item \textbf{GPU:} Experiments are conducted using a server containing four NVIDIA RTX A6000. 
    \item \textbf{RAM:} 1Tb of available RAM.
\end{itemize}

\section{Accuracy Learning Curves}

\begin{figure*}
      \centering
    \includegraphics[width=1\textwidth]{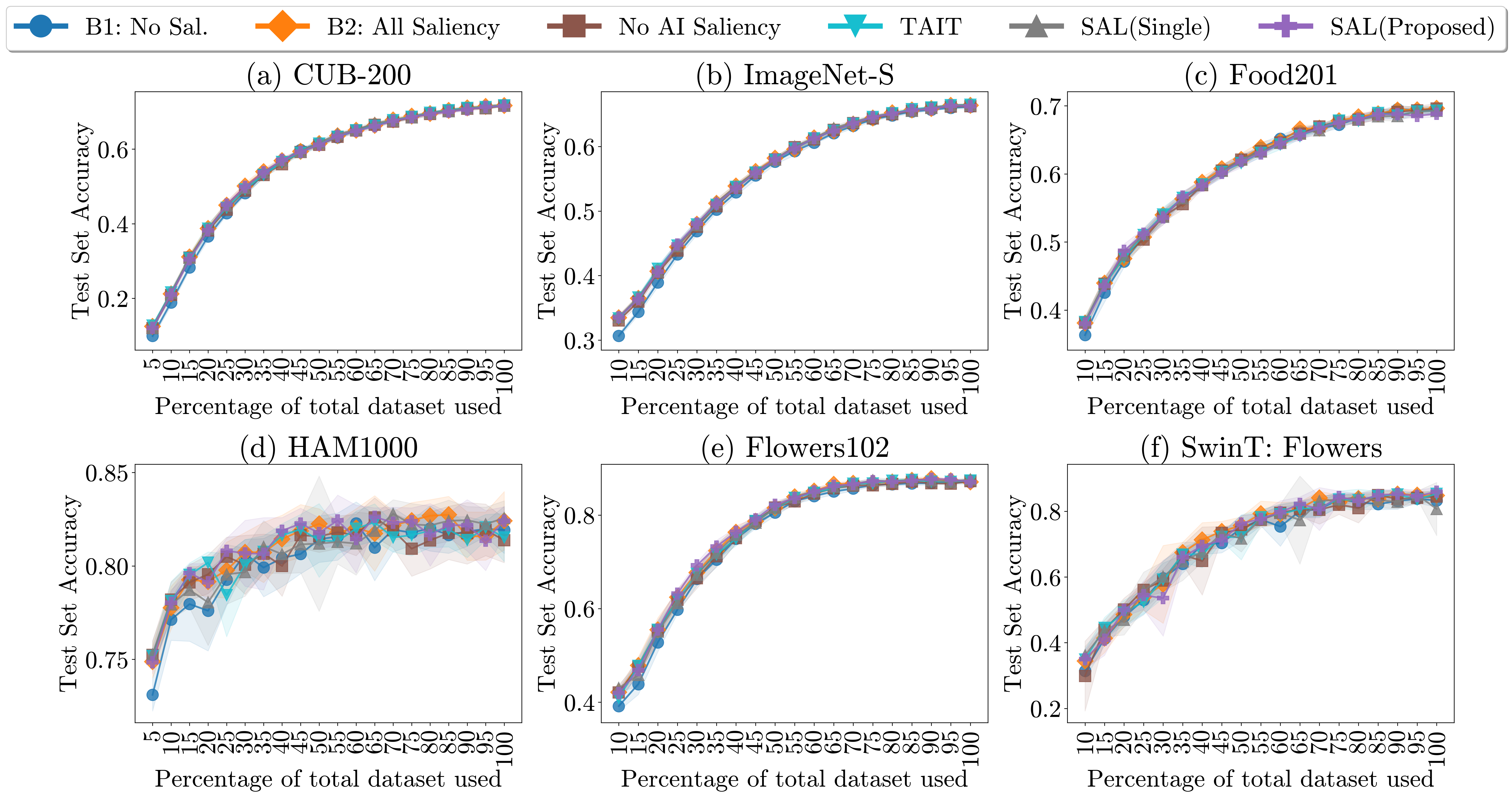}
 \vskip3mm
 \vspace{-1em}
  \caption{Learning curves comparing the accuracy of model saliency trained under various scenarios on the test set for five different datasets. In all cases the AL criteria was margin uncertainty. Plots (a)-(e) are ResNet50-based, while (f) shows the use of SAL with SwinTransformer. Each learning curve shows the mean of 8 AL runs, with the shaded area representing $\pm1\sigma$.}
  \label{fig:test_accuracy}
  \vspace{-1em}
\end{figure*}

Shown in Fig. \ref{fig:test_accuracy} are the associated learning curves for classification performance for the overlap learning curves shown in Fig. 3 in the main text. These plots show that in all studied scenarios, for all studied datasets, the accuracy is similar at all stages. This makes the difference in interpretability more interesting, as we demonstrate that SAL avoids the commonly seen performance/interpretability trade-off.

\section{Learning Curves for SwinTransformer on Other Datasets}

\begin{figure*}
      \centering
    \includegraphics[width=1\textwidth]{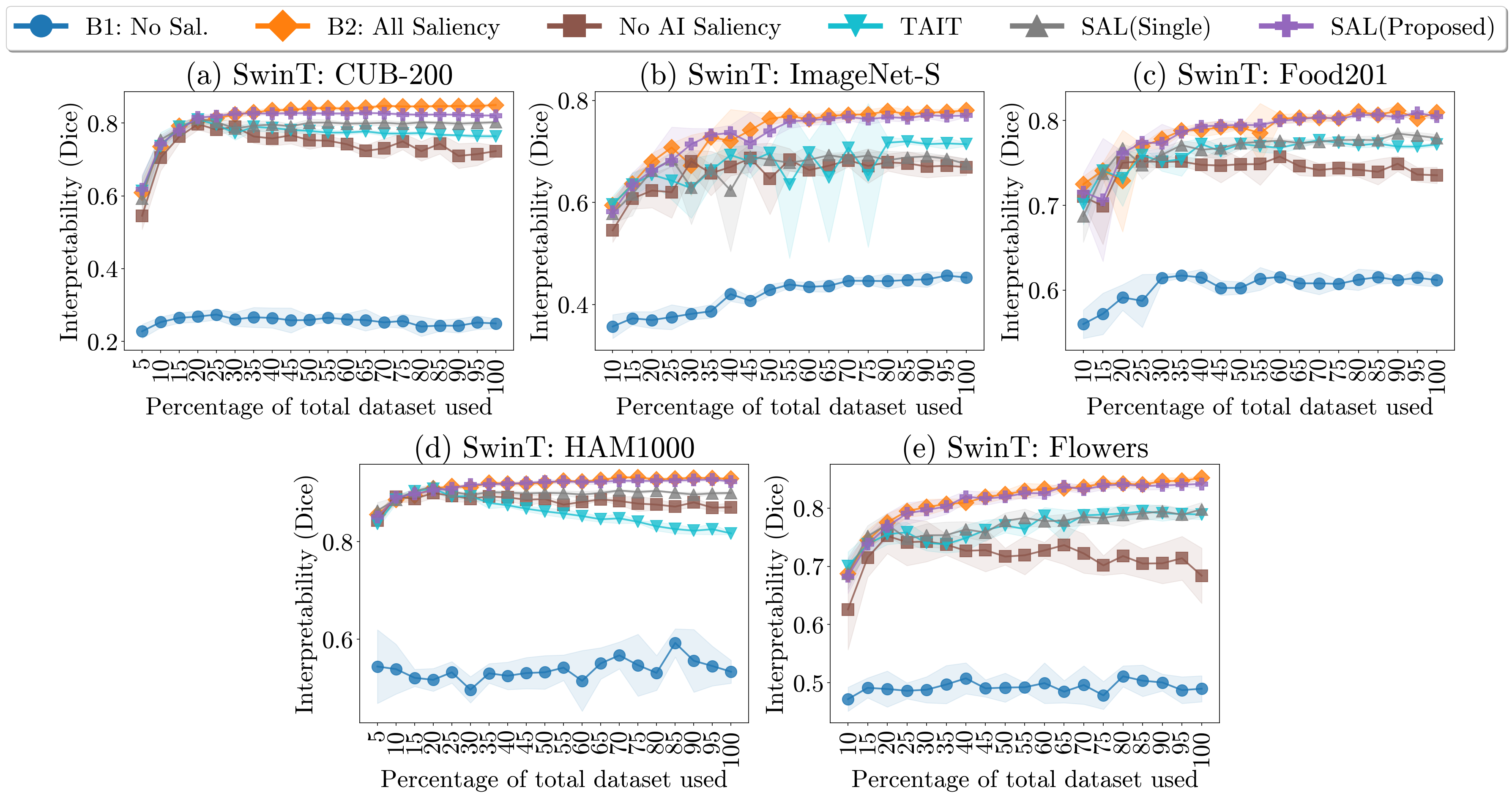}
 \vskip3mm
 \vspace{-1em}
  \caption{Learning curves comparing the overlap/interpretability of model saliency trained under various scenarios on the test set for five different datasets using SwinTransformer. Each learning curve shows the mean of 8 AL runs, with the shaded area representing $\pm1\sigma$.}
  \label{fig:swint-test_overlap}
  \vspace{-1em}
\end{figure*}

\begin{figure*}
      \centering
    \includegraphics[width=1\textwidth]{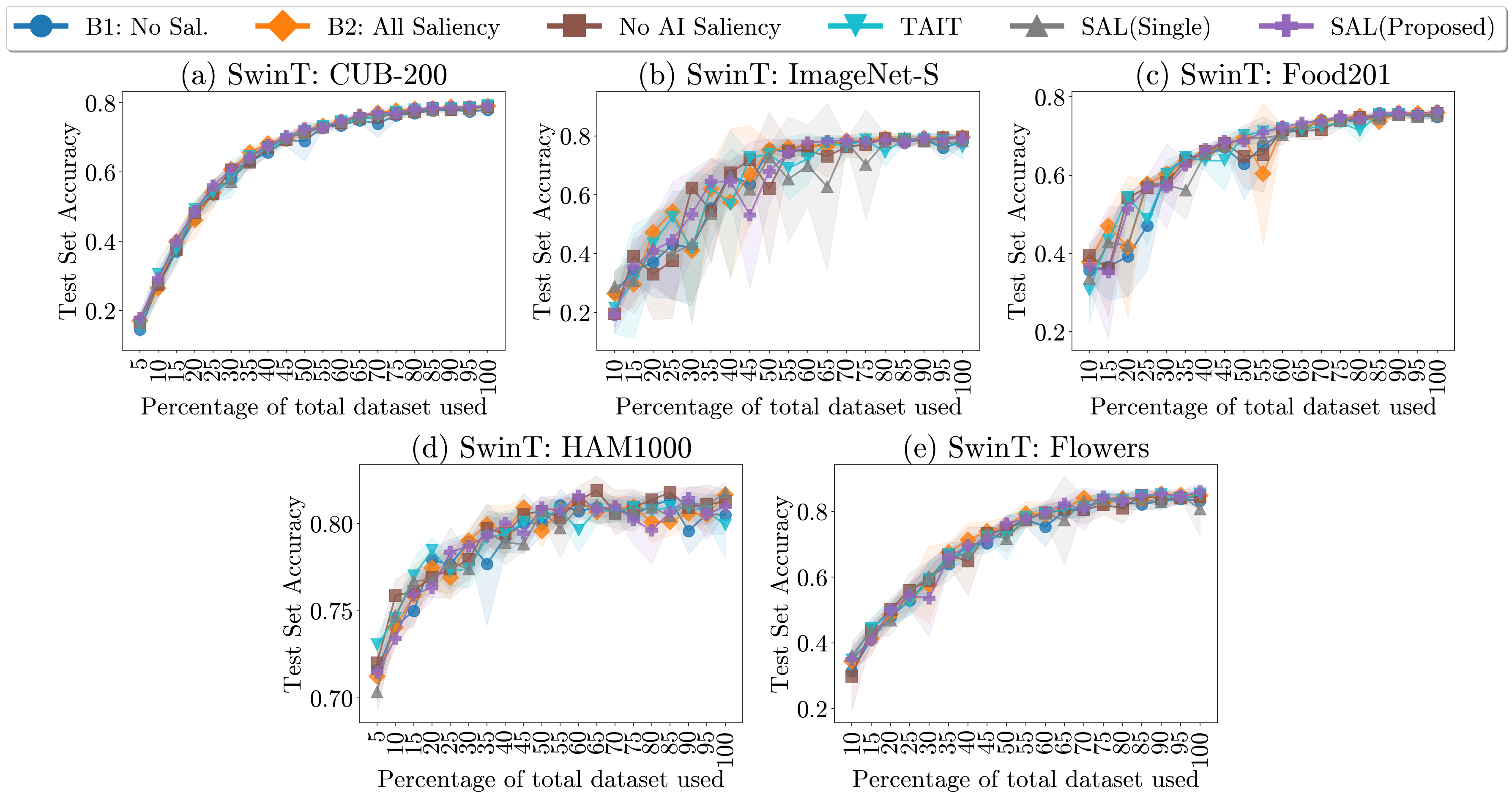}
 \vskip3mm
 \vspace{-1em}
  \caption{Learning curves comparing the accuracy of model saliency trained under various scenarios on the test set for five different datasets using SwinTransformer. Each learning curve shows the mean of 8 AL runs, with the shaded area representing $\pm1\sigma$.}
  \label{fig:swint-test_accuracy}
  \vspace{-1em}
\end{figure*}

In the main text, we just show the learning curves of SwinTransformer~\cite{liu2021swin} on one dataset (Flowers). Here we extend those experiments to detail both the overlap learning curves and accuracy learning curves for SwinTransformer on all five studied datasets. Interpretability learning curves are detailed in Fig. \ref{fig:swint-test_overlap} and accuracy learning curves are detailed in Fig. \ref{fig:swint-test_accuracy}.
As mentioned in the main text, the same trends are apparent when using SAL for both ResNet50 and SwinTransformer.

\section{Using an off-the-shelf Segmenter}
\label{sec:ots}

We run an experiment using the Segment Anything Model \cite{Kirillov_2023_ICCV} in place of human saliency. We selected the output mask with the highest \textit{predicted\_iou} as the segmentation. Results, shown in Fig. \ref{fig:ots-test_overlap} show that SAM generated saliency is preferable to no saliency incorporation, but using SAL with an initial set of human annotations is better. As with human saliency, accuracy is not impacted when using SAL, even with off-the-shelf-segmentations, as shown in Fig. \ref{fig:ots-test_accuracy}.

\section{Changing to AI generated Saliency at different points}

\begin{figure*}
      \centering
    \includegraphics[width=1\textwidth]{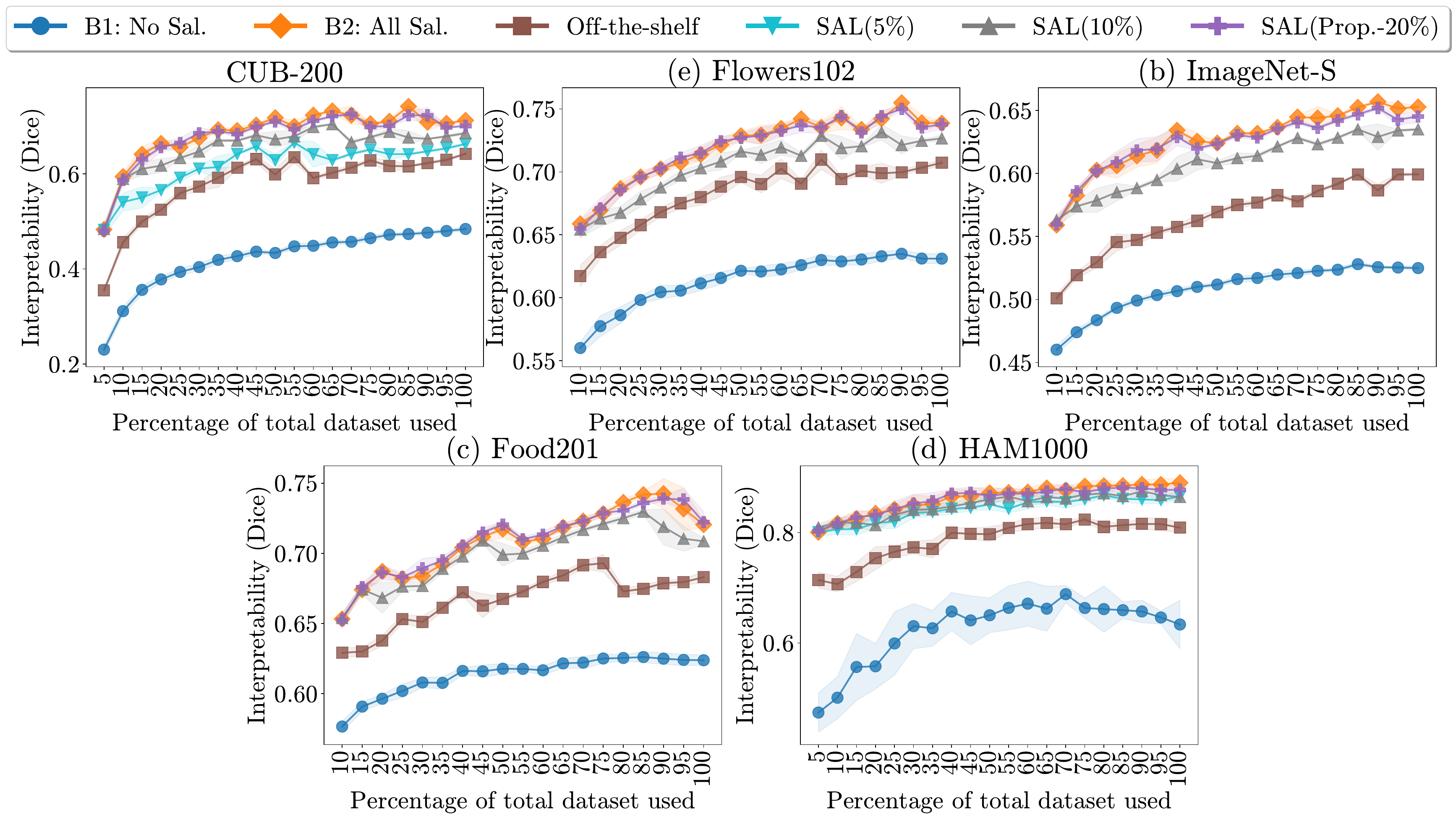}
 \vskip3mm
 \vspace{-1em}
  \caption{Learning curves comparing the overlap/interpretability of model saliency trained under various scenarios on the test set for five different datasets using a ResNet50 backbone. Each learning curve shows the mean of 8 AL runs, with the shaded area representing $\pm1\sigma$.}
  \label{fig:ots-test_overlap}
  \vspace{-1em}
\end{figure*}

\begin{figure*}
      \centering
    \includegraphics[width=1\textwidth]{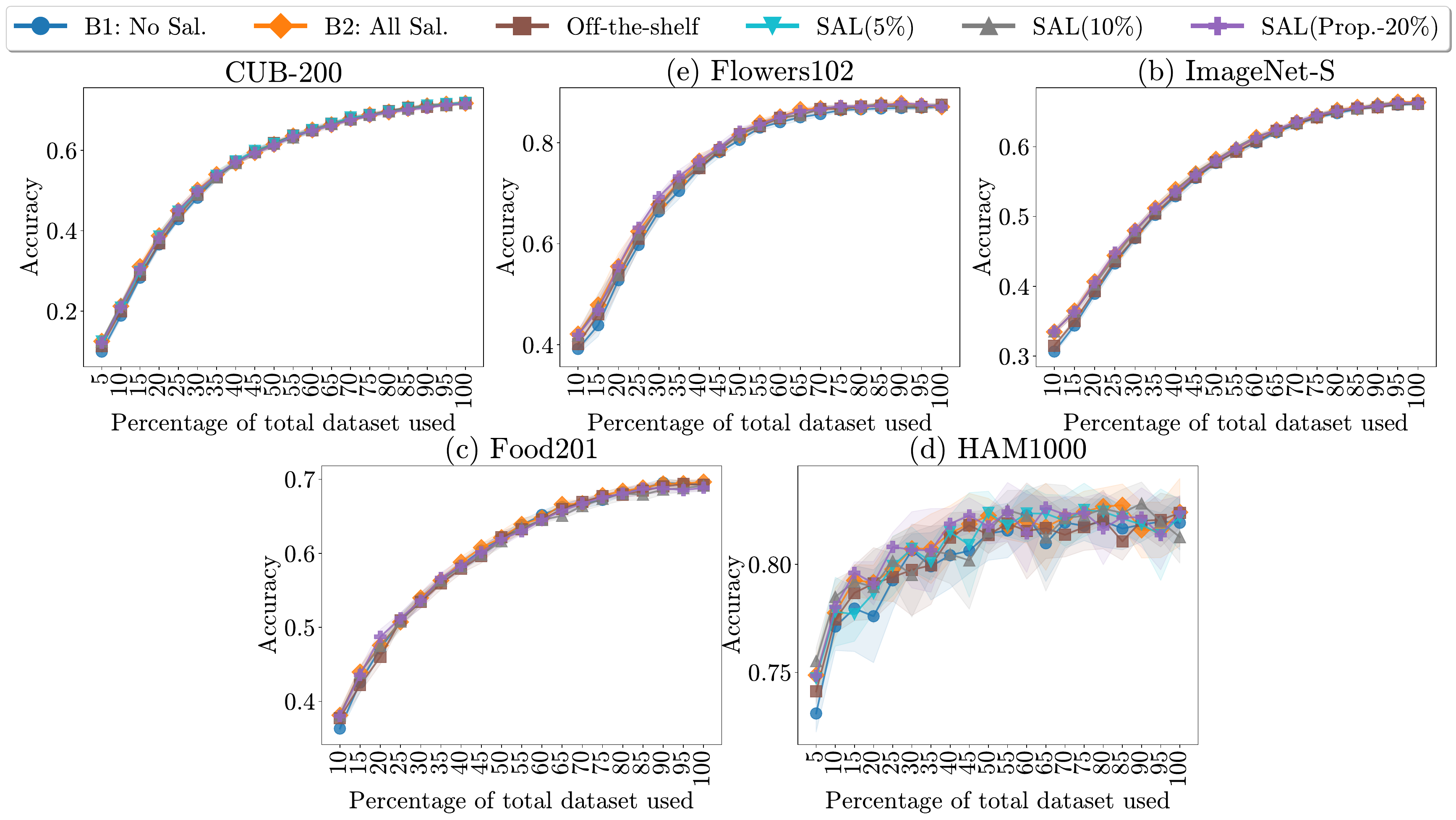}
 \vskip3mm
 \vspace{-1em}
  \caption{Learning curves comparing the accuracy of model saliency trained under various scenarios on the test set for five different datasets. Each learning curve shows the mean of 8 AL runs, with the shaded area representing $\pm1\sigma$.}
  \label{fig:ots-test_accuracy}
  \vspace{-1em}
\end{figure*}

We ran two additional experiments on the all datasets; collecting 5\% human annotations and 10\% human annotations instead of the 20\% used in the main paper. Results in Fig. \ref{fig:ots-test_overlap} show SAL is effective with just 5\% human annotations, and performance increases with more human annotations. SAL is more effective with just 5\% human annotations than using off-the-shelf segmentation. Note that we could not complete experiments using 5\% human annotations for Flowers102, ImageNet-S or Food201 as the minimum initial set needs one annotation per class, and these datasets have only 10 images per class in the training set.

\section{Changing Saliency Probing Method}

In this section we replicate the interpretability learning curves for all datasets using various saliency probing methods. The purpose of this experiment is to show that the performance of SAL is invariant of the saliency probing method. In each experiment using SAL, the generated saliency is created using the specified probing method. Additionally, for all experiments the saliency probing method to evaluate on the test set is also the specified probing method. In all cases for this demonstration, the model architecture used is ResNet50. 

\subsection{GradCAM \cite{gradcam}}

\begin{figure*}[t]
      \centering
    \includegraphics[width=1\textwidth]{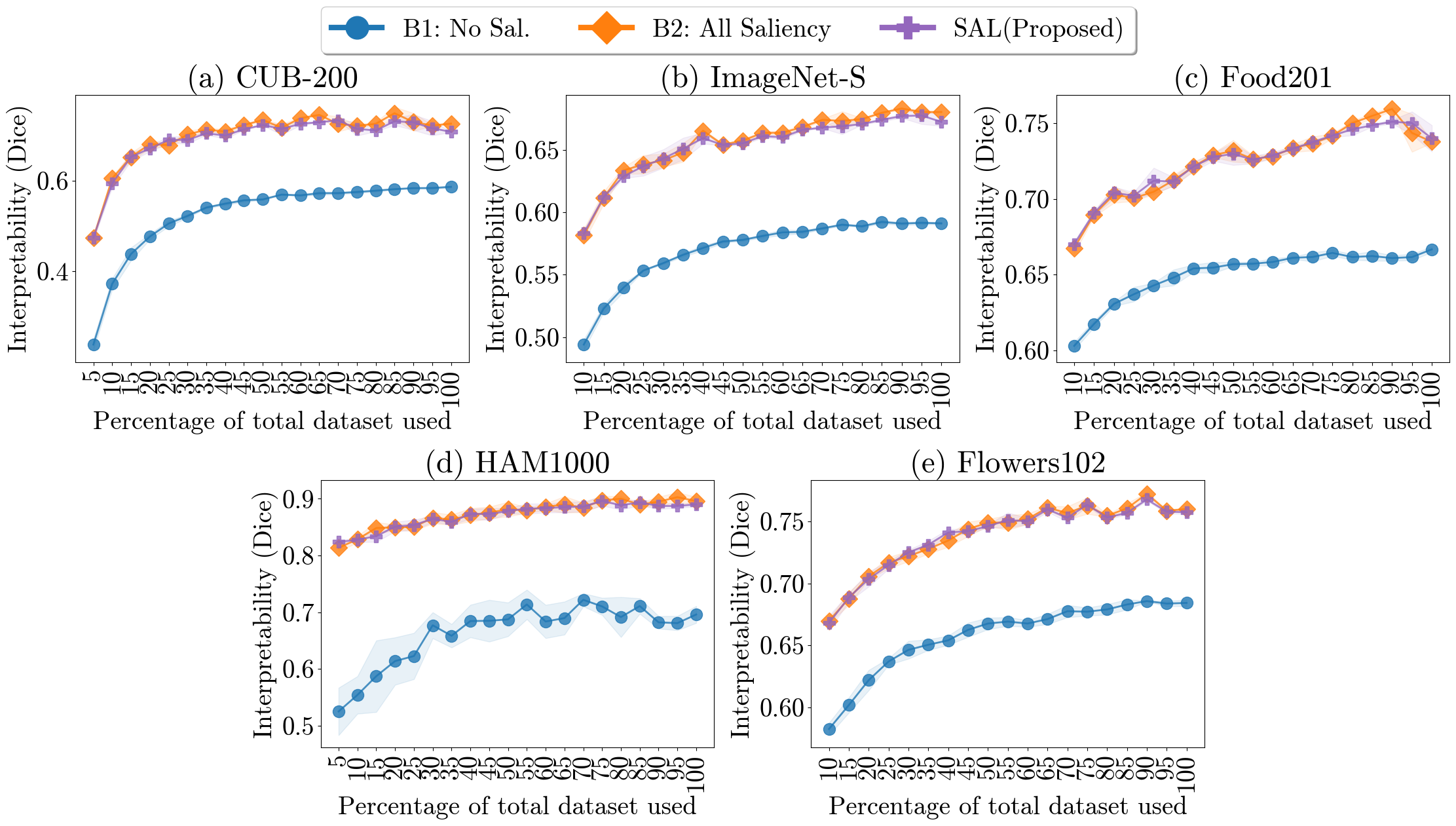}
 \vskip3mm
 \vspace{-1em}
  \caption{Learning curves comparing the overlap/interpretability of model saliency when the saliency probing method employed was \textbf{GradCAM\cite{gradcam}} for five different datasets. In all cases the AL criteria was margin uncertainty with the ResNet50 architecture. Each learning curve shows the mean of 4 AL runs, with the shaded area representing $\pm1\sigma$.}
  \label{fig:gradcam_overlap}
  \vspace{-1em}
\end{figure*}

The results on the test set for five datasets when the saliency probing method is set to GradCAM can be found in Fig. \ref{fig:gradcam_overlap}. As with CAM in the main text, SAL effectively replicates the performance of a fully supervised approach.

\subsection{GradCAM++ \cite{gradcam++}}

\begin{figure*}
      \centering
    \includegraphics[width=1\textwidth]{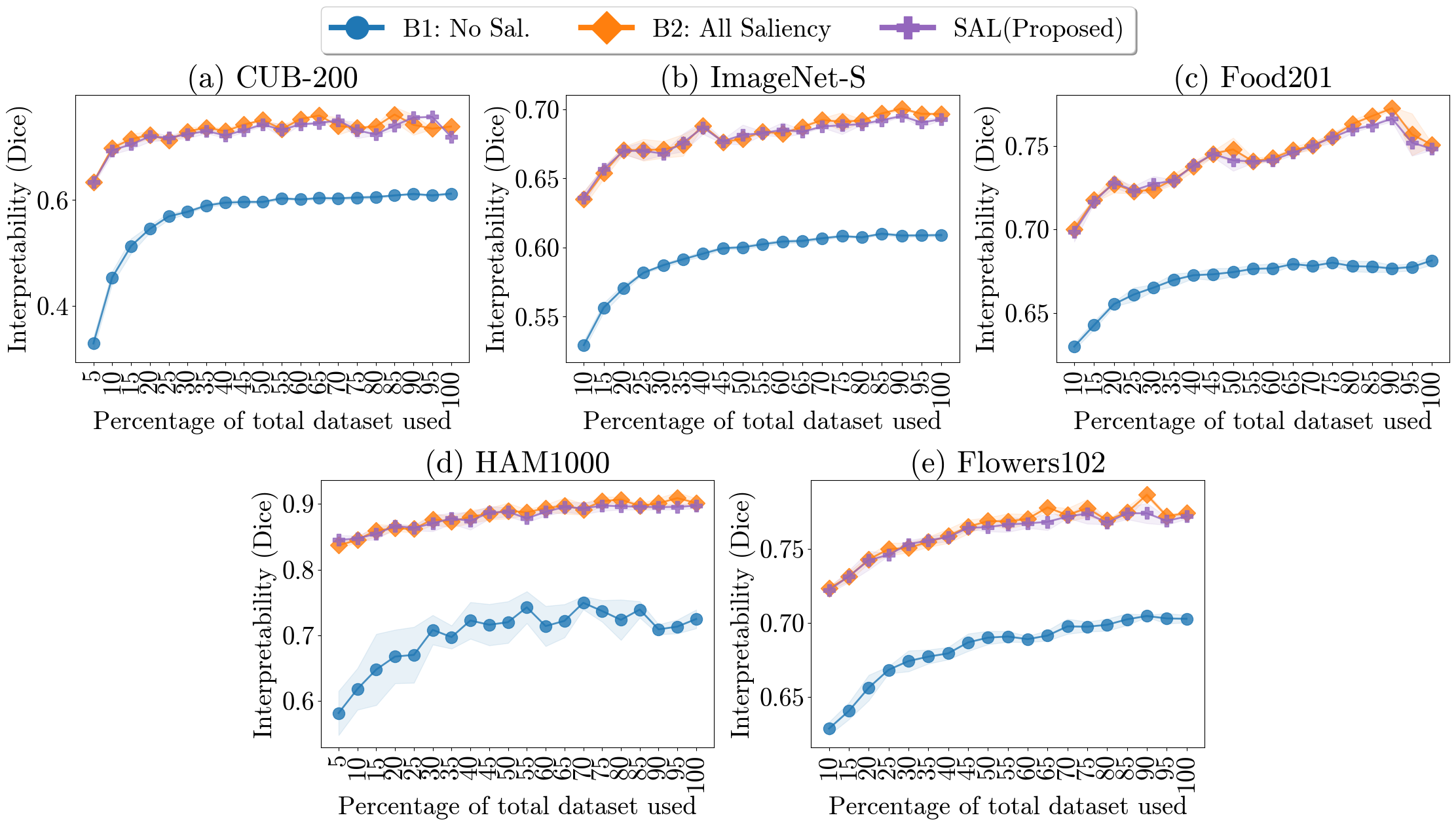}
 \vskip3mm
 \vspace{-1em}
  \caption{Learning curves comparing the overlap/interpretability of model saliency when the saliency probing method employed was \textbf{GradCAM++\cite{gradcam++}} for five different datasets. In all cases the AL criteria was margin uncertainty with the ResNet50 architecture. Each learning curve shows the mean of 4 AL runs, with the shaded area representing $\pm1\sigma$.}
  \label{fig:gradcampp_overlap}
  \vspace{-1em}
\end{figure*}

The results on the test set for five datasets when the saliency probing method is set to GradCAM++ can be found in Fig. \ref{fig:gradcampp_overlap}. As with CAM in the main text, SAL effectively replicates the performance of a fully supervised approach.

\subsection{HiResCAM \cite{hirescam}}

\begin{figure*}
      \centering
    \includegraphics[width=1\textwidth]{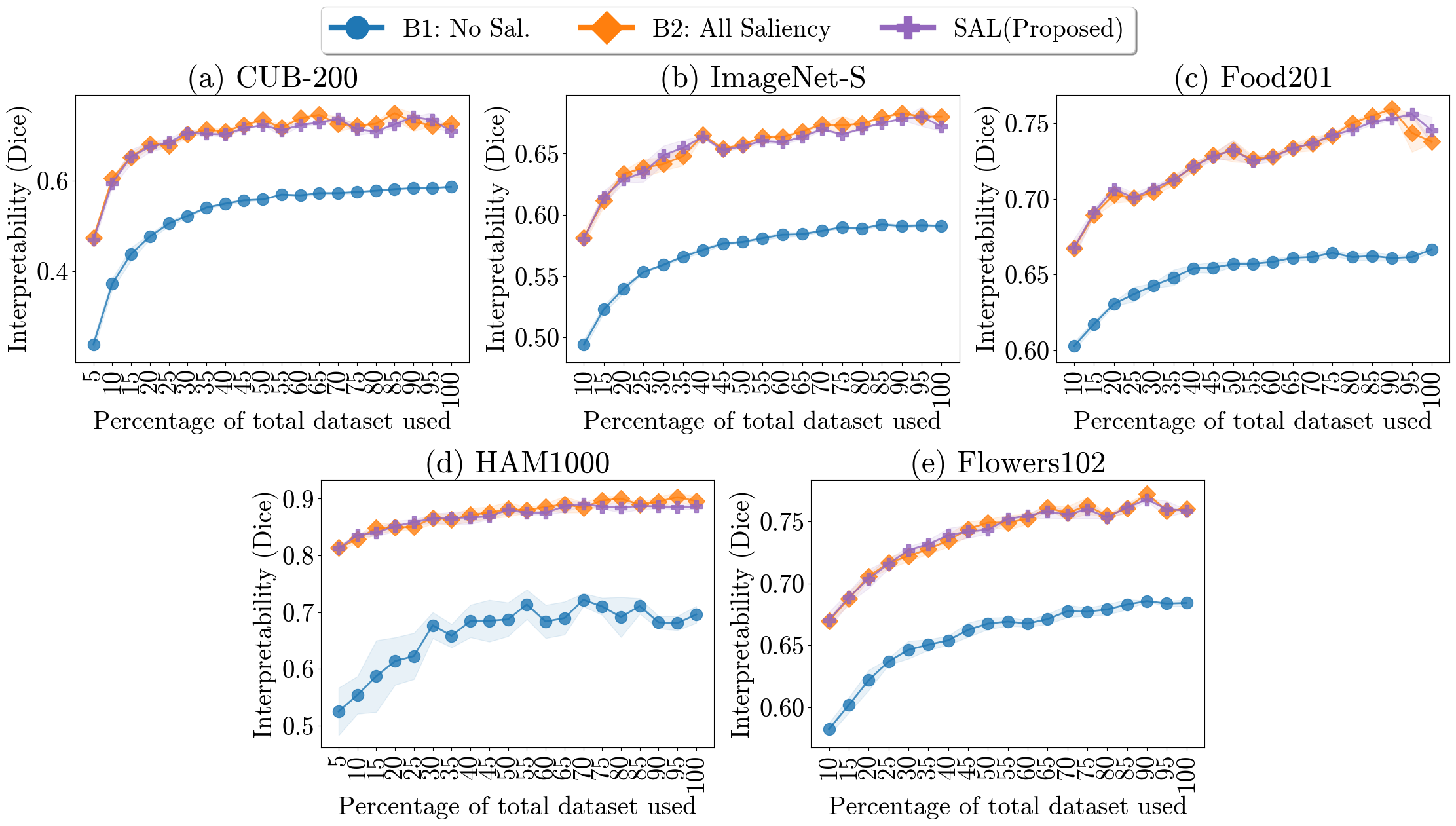}
 \vskip3mm
 \vspace{-1em}
  \caption{Learning curves comparing the overlap/interpretability of model saliency when the saliency probing method employed was \textbf{HiResCAM\cite{hirescam}} for five different datasets. In all cases the AL criteria was margin uncertainty with the ResNet50 architecture. 
  Each learning curve shows the mean of 4 AL runs, with the shaded area representing $\pm1\sigma$.
  }
  \label{fig:hirescam_overlap}
  \vspace{-1em}
\end{figure*}

The results on the test set for five datasets when the saliency probing method is set to HiResCAM can be found in Fig. \ref{fig:hirescam_overlap}. As with CAM in the main text, SAL effectively replicates the performance of a fully supervised approach.

\section{Varying alpha parameter in CYBORG}

\begin{figure*}
  \begin{subfigure}[b]{1\textwidth}
  \centering
      \begin{subfigure}[b]{0.42\textwidth}
          \centering
          \includegraphics[width=1\textwidth]{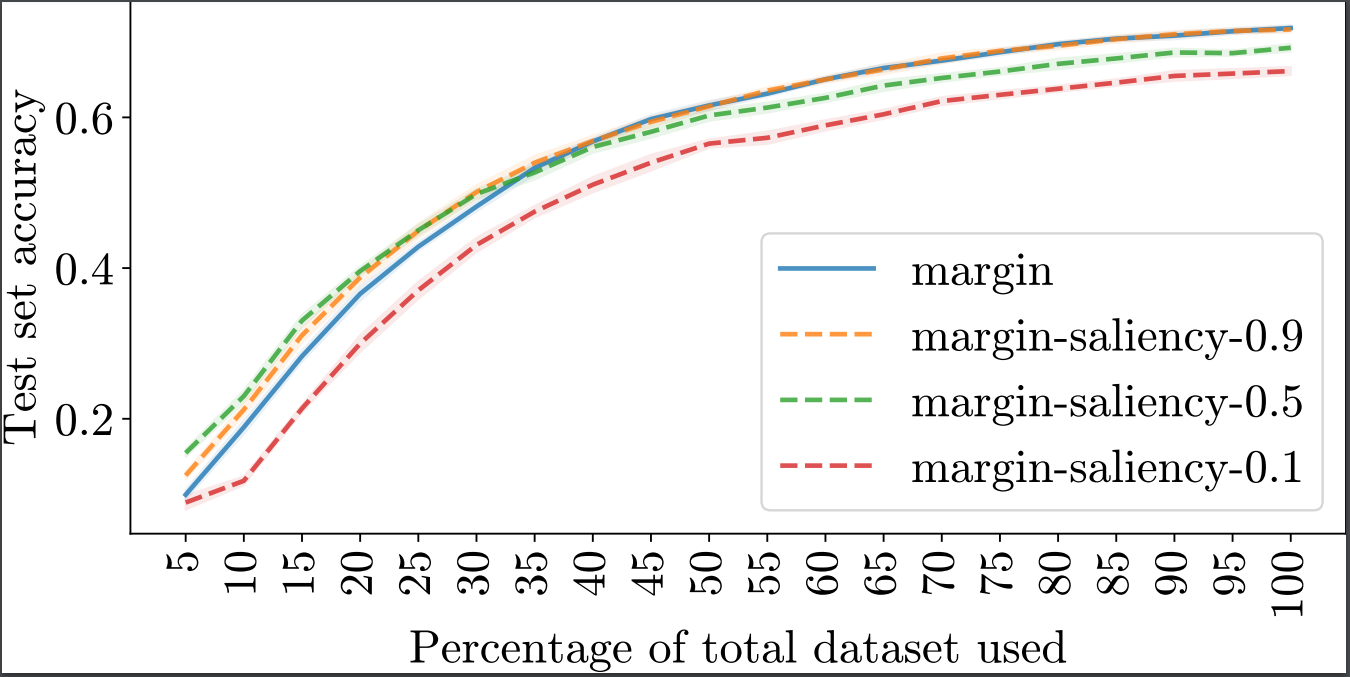}
          \caption{Accuracy}
      \end{subfigure}
      \begin{subfigure}[b]{0.42\textwidth}
          \centering
          \includegraphics[width=1\textwidth]{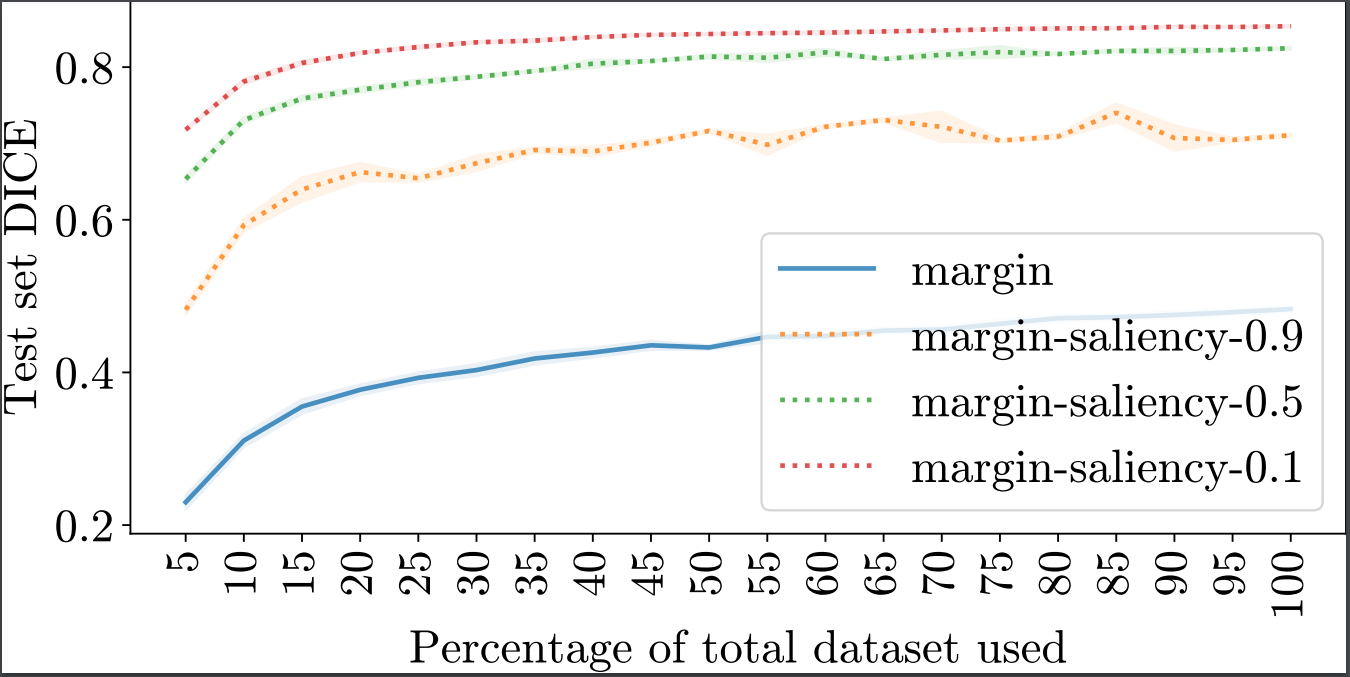}
          \caption{Overlap/Interpretability}
      \end{subfigure}
   
  \end{subfigure} \vskip3mm
  \caption{Figure detailing the classification performance/interpretability trade-off. As the $\alpha$ parameter changes, the emphases is moved between predicting the saliency and predicting the class label. This variation provided the inspiration for SAL.}
  \label{fig:varying_alpha}
\end{figure*}

As seen in the definition of CYBORG loss \cite{boyd2021cyborg} from the main text (Sec. 3.1), the $\alpha$ parameter controls the balance between classification performance and focus on the saliency. In \cite{boyd2021cyborg}, the authors set this value to $\alpha = 0.5$, equally balancing both components. Fig. \ref{fig:varying_alpha} shows the effect on both (a) accuracy and (b) interpretability when the alpha value is adjusted for the Birds dataset. The performance/interpretability trade-off is evident in this figure. Alpha values closer to 0 put more focus on learning the saliency, at the direct expense of accuracy. Alpha values closer to 1 show significantly better interpretability, with a large decrease in accuracy. This imbalance motivated the design of the solution in this work. In SAL, two models are trained: one for high interpretability and one for high accuracy.


\end{document}